\documentclass[times, 10pt,twocolumn]{article}
\usepackage{latex8}
\usepackage{times}
\usepackage{balance}
\usepackage{epsfig}
\usepackage{amsmath}
\usepackage{amssymb}
\newcommand{\comment}[1]{}
\begin{document}

\title{A Medial Axis Based Thinning Strategy for Character Images}



 \author{Soumen Bag, Gaurav Harit\\
 \emph{Department of Computer Science and Engineering}\\
 \emph{Indian Institute of Technology Kharagpur, Kharagpur-721 302, India}\\
 \emph{\{soumen, gharit\}@cse.iitkgp.ernet.in}\\
 }

\maketitle
\thispagestyle{empty}

\begin{abstract}

Thinning of character images is a big challenge. Removal of strokes or deformities in thinning is a difficult problem.
In this paper, we have proposed a medial axis based thinning strategy used for performing skeletonization of printed and
handwritten character images.
In this method, we have used shape characteristics of text to get skeleton of nearly same as the true character shape.
This approach helps to preserve the local features and true shape of the character images.
The proposed algorithm produces one pixel width thin skeleton. As a by-product of our thinning approach, the skeleton also
gets segmented into strokes in vector form. Hence further stroke segmentation is not required.
Experiment is done on printed English and Bengali characters and we obtain less spurious branches comparing with other thinning
methods without any post processing.

\end{abstract}

\Section{Introduction}

Thinning of shape has a wide range of application in image processing, machine vision, and pattern recognition. But removal of
spurious strokes or shape deformation in thinning is a difficult problem.
In the past several decades many thinning algorithms have been developed considering all these problems \cite{Lam11}.
They are broadly classified into two groups: raster scan based and medial axis based. Raster scan based methods are classified
into two other categories: sequential and parallel.
Sequential algorithms consider one pixel at a time and visit the pixel by raster scanning \cite{Baja16} or contour following \cite{Arcelli12}.
Parallel thinning algorithms are based on iterative processing and they consider a pixel for removal based on the results of previous iteration
only \cite{Datta13,Huang01,Leung02,Zhang14,Zhu03}. Many of the raster scan based character thinning methods can not preserve the local properties
or features of the character images properly. As a result, they give slightly shape distorted output.
Medial axis based methods generate a central or median line of pattern directly in one pass without examining all the individual pixels \cite{Perez15}.
They also give slight distorted result at some local regions.
Here we attempt to minimize local distortions by making use of shape characteristics of text. Next we begin a brief description about few prominent
raster scan based thinning algorithms.

Datta and Parui~\cite{Datta13} have proposed a parallel thinning
algorithm which preserves connectivity and produces skeleton of one
pixel thickness. Each iteration of the algorithm is divided into four sub-iterations. These sub-iterations use two 1 x 3 and two 3 x 1 templates for
 removing boundary pixels: east, west, north, and south respectively. The method uses one 3 x 3 window to avoid the removal of critical point
 (which alters the connectivity) and end point (which shortens a leg of the skeleton).

Leung {\em et al.}~\cite{Leung02} have introduced a contour following thinning method having no sub-iteration. They have used a lookup table to
 avoid the use of multiple templates for removing boundary pixel. The lookup table has 256 entries and each entry contains three fields
 (neighbor number, weight number, and connection number). The number of entries depends on the different possibilities of 8-connectivity of a pixel.
  The determination of the possibility of pixel removal is depends on the values of  the patterns of the spatial contour pixels.

To improve the pixel connectivity of thinned character images, Huang
{\em et al.}~\cite{Huang01}, have introduced a new set of templates
(three  4 x 3, one 4 x 4, and three 3 x 4). This algorithm considers
all possible patterns of 8-connectivity of a pixel (similar
to~\cite{Leung02}) and generates template based elimination rule for
deleting boundary pixels. The pixel deletion is done based on the
number of black pixels in the
 8-neighbor connectivity.  Additionally, it compensates information loss by integrating the contour and skeleton of pattern. The information loss
 is detected based on the ratio of the skeleton and contour pixels. If the value is less than a predefined threshold then the thinned image is
 replaced by the contour image.

The main contribution of Zhu's method ~\cite{Zhu03} is that it does not consider all pixels equally for performing character thinning. It is based
on the substitution of pixels of strokes or curves which are most valuable part for character recognition. This method uses a set of thinning
templates, similar to~\cite{Datta13}, for boundary pixel removal and handling corners or junction points properly. This method has tried to
preserve the shape of character images such as Chinese characters, English alphabets, and numerals after thinning. All the above discussed
algorithms perform boundary pixel deletion using various thinning templates or values of lookup table. Now we are discussing a medial axis
based thinning algorithm.

For performing thinning of regular shape, Martinez-Perez {\em et al.}~\cite{Perez15} have introduced a medial axis based thinning method.
This approach does not use thinning templates or lookup table for performing thinning. They have generated medial points in between two parallel
 contour segments and repeat the procedure for all remaining parallel contour segments. But they have not applied this concept for curve.
 Finally, we give a brief description of a thinning method which is used for skeleton simplification.

Telea {\em et al.}~\cite{Telea04} have introduced a method to simplify skeleton structure of an image by removing few skeleton branches.
The simplification is done by analyzing the quasi-stable points of the Bayesian energy function, parameterized by boundary of contour and
internal structure of the skeleton.  The experimental results show that it gives multi-scale skeleton at various abstract levels.

All the above mentioned algorithms can not retain the shape of the character images properly. From that point of view, we have addressed a
medial axis based thinning method for generating proper skeleton of character images. The algorithm is non iterative and template free. The
 main advantage is that it uses shape characteristics of text to determine the areas within the image region to stop thinning partially. This
  approach helps to preserve the local features and true shape of the character. Additionally, it produces a set of vectorized \textit{strokes}
   with the thin skeleton as by-product. These resultant skeletons have stronger ability to oppose shape deformation and more convenience to
   feature extraction and classification for OCR.

This paper is organized as follows. Section 2 describes the methodology of the proposed algorithm. Section 3 contains the experimental results
 and there comparison with other thinning methods. This paper concludes with some remarks on the proposed method and future work in section 4.

\Section{Medial axis based thinning algorithm}

In this section we describe our proposed thinning strategy based on
contour extraction and medial axis extraction for text alphabets.
The advantages of our proposed technique are:

        \begin{itemize}
            \item   It provides a vectorized output of the thin skeleton.
            \item   The vectorized output is a collection of strokes.
                    Hence further stroke segmentation is not required.
            \item   Many of the spurious thinning branches which inevitably occur when applying other rasterized thinning algorithms do not occur
             in our proposed algorithm.
            \item   Our approach is most suitable for thinning text alphabets, printed or handwritten. Particularly it gives correct output even
            in the presence of changing width of the strokes.
        \end{itemize}

\SubSection{Preprocessing and contour extraction}

            Given a scanned document page we binarize it using the Otsu's algorithm~\cite{gonzalez-woods-02}.
            Currently we are working with documents with all text content, hence we identify the alphabets
            as connected components in binary images.
            This works for English alphabets, however, for Bengali or Hindi documents the entire word gets
            identified as a single connected component because of the \textit{matra/shiro-rekha} which connects the individual characters.
            For this case we separate out the individual \textit{akshara} within a word by making use of vertical profile, and delineating
            adjacent \textit{akshara} within a word at the minima points of the profile.

            Given an isolated alphabet, its boundary contour is extracted.
            We detect a boundary pixel on the alphabet using a $3\times 3$ mask centered on every black pixel of the character.
            If there is even a single background (white) pixel within the mask it would
            imply that the center black pixel is flagged as a boundary pixel.
            Starting from this first boundary pixel we traverse the boundary using a connected component aggregation
            algorithm, i.e. we go on aggregating the black pixels which are adjacent and also on the boundary.
            This produces a vector representation of the boundary contour.
            This boundary contour is then segmented into straight line or curved segments which we
            refer to as \textit{contour strokes}.

\SubSection{Contour segmentation}
            In this section we describe our algorithm to convert the vectorized boundary contour into component
            segments $-$ straight lines or curves.
            The novelty of this algorithm lies in its \textit{search}-based identification of
            contour segmentation points.
            Given a vectorized contour the candidate segmentation points are identified by analyzing the
            incremental changes in the orientation as pixels are added at one end of a hypothesized contour
            segment.
            The orientation of a contour segment is the slope of the line joining its hypothesized end points.
            A hypothesized contour segment initiated from a \textit{start point} $s$ is denoted $C_s$.
            The other end i.e. the \textit{end point} is allowed to extend to the adjacent contour pixel $x_1$, then to the next one $x_2$
            in sequence and so on.
            Our objective is to find a suitable contour boundary pixel for the end point of $C_s$.
            For each possible \textit{end point} location $e_i$ we get a hypothesized contour segment $\widehat{s e_i}$.
            The orientation $\theta_i$ of this segment is the slope of the line joining the points $s$ and $e_i$.
            We analyze the change $\Delta\theta_i = \theta_i - \theta_{i-1}$.
            For a sequence of hypothesized end point positions $e_i = \dots, x_{k-1}, x_k, x_{k+1}, \dots$ we
            note the trend of $\Delta\theta_i$ being positive,
            negative, or zero.
            We form groups of successive pixels for which $\Delta\theta_i$ remains positive, negative, or zero.
            In other words the pixels forming a group are successive and have a common trend for  $\Delta\theta_i$ .
            For each such group, say $G_p$, which spans pixels, say $x_{p1}$ to $x_{p2}$, we compute the total
            change in orientation, $\Delta\theta_{Gp} = \theta_{p2} - \theta_{p1}$.
            We also note the total number of pixels $N_{Gp}$ in the group $G_p$.

            Since our objective is to identify a candidate end point $e$ for the hypothesized contour segment $C_s$
            which attempt to choose a suitable position for $e$
            from the available end points of the pixel groups $G_p$'s which have been identified with respect to the
            start point $s$ of the contour segment $C_s$.

        \begin{itemize}
            \item   If $\Delta\theta_{Gp} > \boldsymbol{\theta}_{\text{threshold}}$ (value is set to 0.7) for any group $G_p$, then take
             the end point of $C_s$ as the end
                    point of $G_{p-1}$.
                    If $G_p$ is the first pixel group from the start point $s$, then restrict $G_p$ to extend up to only that pixel
                    where $\Delta\theta_{Gp} = \boldsymbol{\theta}_{\text{threshold}}$.

            \item   If $N_{Gp} > \boldsymbol{N}_{\text{threshold}}$  (value is set to 5) for any group $G_p$,
                    then count the total number of pixels $N_{s}$ in $C_s$ up to
                    the end pixel of $G_{p-1}$.
                    If $N_s > \boldsymbol{N}_{\text{threshold}}$ then take the   end
                    point of $G_{p-1}$ as the end point of $C_s$.
        \end{itemize}

                In the above formulation $G_{p-1}$ denotes the group just preceding $G_p$.
                Once we get the candidate end point for the hypothesized contour segment $C_s$,
                we do a refinement of its location on the contour.
                In other words we search for its most appropriate position in the vicinity of the
                identified candidate position.
                This search is based on analyzing local variance of orientations in the forward
                and backward direction from each contour pixel position in the vicinity of the candidate end point.
                The position where both the forward and the backward variances are minimum/low is taken as
                the most appropriate position of the end point of $C_s$.

\begin{figure}[h]
                    \begin{center}
                    \includegraphics[width=.24\textwidth,height=.13\textheight]{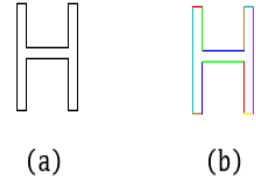}
                    \caption{(a) Contour image  (b) Contour strokes (see in color)}\label{fig:illustrativeEg}
                    \end{center}
                    \end{figure}

                The result of contour segmentation for Fig.~\ref{fig:illustrativeEg}(a) is shown in
                Fig.~\ref{fig:illustrativeEg}(b) where the identified contour strokes are shown in a
                different color.
                Each contour stroke corresponds to a straight line stroke or a curved stroke within
                the character. (Note that our notion of a stroke is different from the one used in handwriting recognition).
                Given that the character boundary contour has been now segmented into \text{contour-strokes},
                as the next step we get the
                medial axis for the character.

\SubSection{Getting the medial axis}

                We obtain the medial axis using the pixels on the boundary contour partitioned into contour strokes.
                To obtain the medial axis we need to identify the parallel contour-strokes on the boundary contour.
                While doing so we incorporate the most obvious text-specific knowledge: the distance between the parallel contour-strokes
                 should be small (since the pen-width is generally small), and all the pixels in-between the two parallel contour-strokes
                 should be black i.e. belong to the character itself and not the background.

                Processing starts from any contour-stroke, say $C_s$.
                We process the pixels $\{x_1,x_2,...x_{Ns} \}$ on the contour segment $C_s$.

        \begin{enumerate}
            \item   For a pixel $x_i$ on $C_s$ compute the local orientation.
                \begin{eqnarray}
                  \theta_{\text{local}^{x_i}} = \frac{1}{2} \left[ \underset{j=1~\text{to}~5}{\text{Avg}}\!\! {\theta(x_i,x_{i+j})}\!+  \!\!\!\!\underset{j=1~\text{to}~5}{\text{Avg}}\!\! {\theta(x_i,x_{i-j})}\right] 
             \nonumber
             \end{eqnarray}        
                        where $\theta(x_i,x_{i+j})$ refers to the orientation of the line segment joining pixel $x_i$ to
                        the $j^{\text{th}}$ pixel following down the contour (i.e. forward direction),
                        and $\theta(x_i,x_{i-j})$  refers to the orientation of the line segment joining pixel $x_i$ to
                        the $j^{\text{th}}$ pixel preceding up the contour (i.e. backward direction).
                        The averaging operation is denoted by $\text{Avg}$.

            \item   Compute the direction $\perp^{x_i}$ perpendicular to the local orientation  $\theta_{\text{local}^{x_i}}$.
            \item   Starting from pixel $x_i$ on $C_s$, traverse pixels along the direction of
                            $\perp^{x_i}$ till a border pixel (not on contour $C_s$) is reached.
                            Let this border pixel be denoted as $x'_i$.
            \item   Compute the local orientation $\theta_{\text{local}^{x'_i}}$ at the pixel $x'_i$.
            \item   Compute the distance $d_{x_ix'_i}$ between the points $x_i$ and $x'_i$.
            \item   If $\theta_{\text{local}^{x_i}}$ is nearly same as $\theta_{\text{local}^{x'_i}}$ ( $\left|\theta_{\text{local}^{x_i}}
             -  \theta_{\text{local}^{x'_i}} \right| < 0.45$)
                    and  $d_{x_ix'_i} < \text{\it pen width}$ (value is set to 12), then mark the mid point of the line
                    joining $x_i$ and $x'_i$ as a pixel on the medial axis.
                    If the two conditions are not satisfied then the medial pixel is \textit{not} marked since
                    this is likely to be the junction point of two or more pen strokes, hence the ambiguity
                    has to be resolved later.

            \item   Flag off the pixels $x_i$ and $x'_i$ as been processed and move to the
                            pixel $x_{i+1}$ and repeat the above steps.
        \end{enumerate}

            The above steps are repeated for all contour strokes.
            For each contour stroke we consider only the pixels which have not been flagged as already processed.

\begin{figure}[h]
                    \begin{center}
                    \includegraphics[width=.12\textwidth,height=.13\textheight]{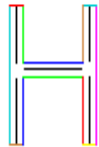}
                    \caption{Medial axis segments (see in color)} 
\label{fig:medialAxisSegments}
                    \end{center}
                    \end{figure}

            The output of this step is given in Fig.~\ref{fig:medialAxisSegments}.
            We see from the result that the medial axis of the character has come up as segments.
            These medial axis segments now need to be extrapolated to join the neighboring segments
            while ensuring that the extrapolation is within the character region.

\SubSection{Extrapolating thin segments in ambiguous regions}

            We consider the medial axis segments of a character as forming the set
            $\mathcal{M} = \{ (M^s_1, M^e_1),  (M^s_2, M^e_2),..., (M^s_m, M^e_m), \}$, where $M^s_i$ and $M^e_i$ denote the start and end points of the $i^{\text{th}}$
            medial axis segment.
            The start/end points now need to be extended so that the neighboring medial axis segments
            can be joined together.
            The steps to identify the neighboring medial axis segments are as follows:
        \begin{enumerate}
            \item   For each start/end point of a medial axis segment find the close start/end points belonging to other medial axis segments.
                    For this purpose we consider that the distance between the two start/end points
                    should be less than a threshold and that the line joining the two start/end points
                    should be within the character region.
            \item   If a given start/end point has two or more start/end points of other segments close enough
                    then we make a proximity set, e.g. $\mathcal{P} = \{ M^s_a, M^s_b, M^e_c  \}$
                    would indicate start/end point of three medial axis segments  $a$, $b$, and $c$.
        \end{enumerate}

            Each proximity set would correspond to start/end points which belong to the junction region
            belonging to multiple pen strokes.
            Identifying the medial axis points in the junction region of pen strokes give
            skeleton points which tend to distort the true shape of the character.
            Our approach avoids identifying the medial axis points in such ambiguous regions and
            instead tries to extrapolate the start/end points of the medial axis segments in a proximity group
            such that the result would be very close to the true shape of the character.
            This is the major contribution of this work.

            The steps for extrapolation are as follows:
        \begin{enumerate}
            \item In a given proximity set $\mathcal{P}$ if any two medial axis segments have the
                    same orientation then any one of the two close start/end points is extended to meet the
                    other one.
            \item   If there is a medial axis segment whose orientation does not match with any other
                    segment in the same proximity set, then its start/end point is extended till it meets the
                    medial axis (extrapolated from other segments) or a junction
                    (formed by extrapolating from other segments in $\mathcal{P}$)
                    on the medial axis or to the start/end point
                    of the other medial axis segment. Of the three cases mentioned which one applies depends on the
                    number of start/end points which are part of a proximity set.
        \end{enumerate}

\begin{figure}[h]
                    \begin{center}
                    \includegraphics[width=.12\textwidth,height=.13\textheight]{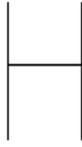}
                    \caption{Skeleton image}\label{fig:skeletonimage}
                    \end{center}
                    \end{figure}

        The result of extrapolation is showed in Fig.~\ref{fig:skeletonimage}.  Note that extrapolation result is almost the true shape of the character and does not produce spurious segments. Our approach gives correct thinning results even for very thick strokes. 

\Section{Experimental results}

At the time of evaluating the performance of thinning algorithm, we
need to take care of the following criteria:
\begin{itemize}
\item The algorithm must preserves the connectivity.
\item The width of resultant skeleton is one pixel.
\item The skeleton edges are not shorten.
\item The skeleton is very close to medial axis.
\item The local properties or features of the character images are preserved after the completion of thinning.
\item The thinning result is very close to the true shape of the input image.
\end{itemize}

We have tested our proposed algorithm on printed English and Bengali character images. All the programs are written in C++ using
Opencv 1.0.0 in the UNIX platform. The \textit{pen width} has to be provided so as to be proportional to the average height of the characters.

\SubSection{Test results of the proposed algorithm}

\begin{figure}
    \begin{center}
    \includegraphics[width=.5\textwidth,height=.9\textheight]{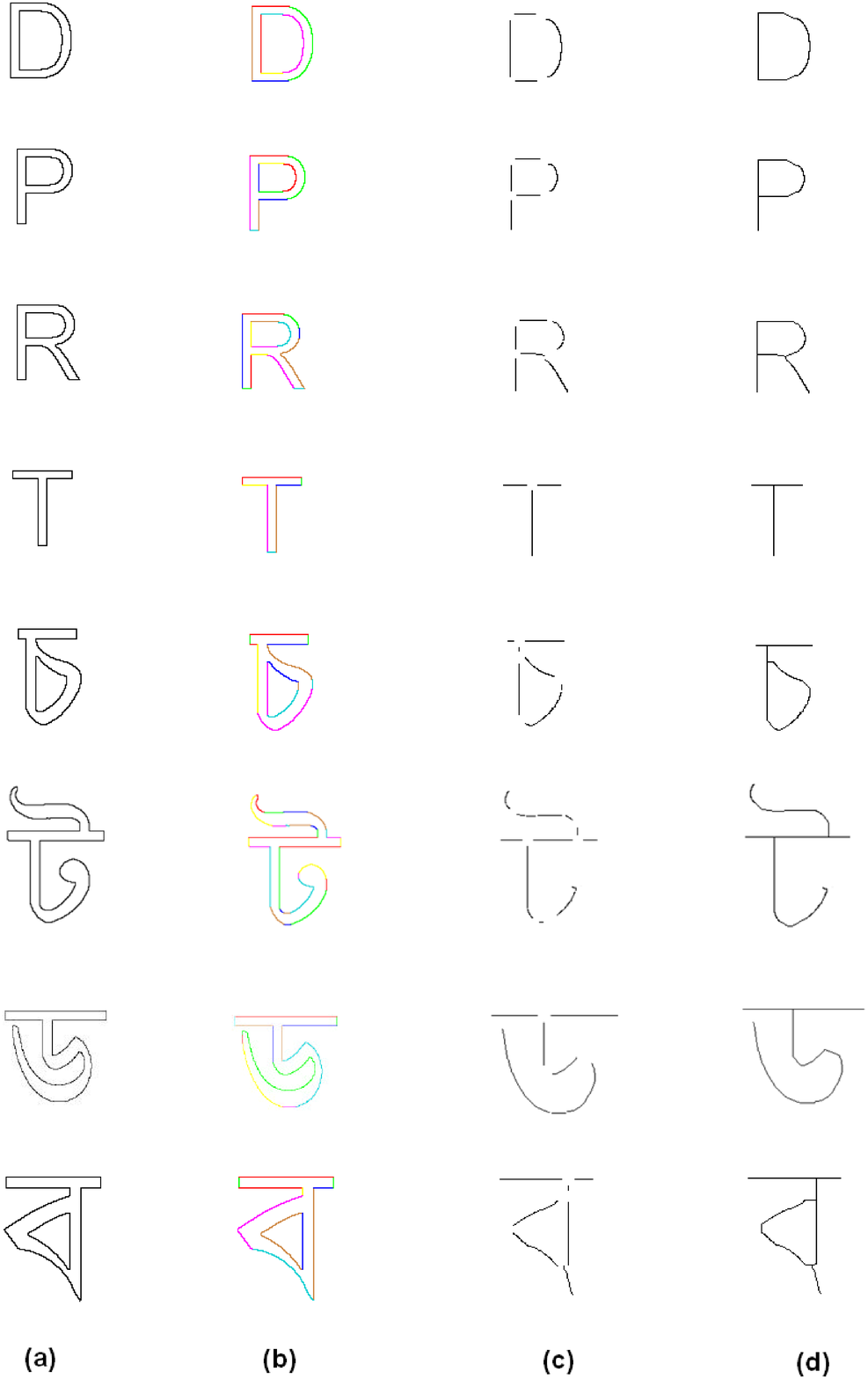}
    \caption{Experimental result (a) Contour image,
            (b) Contour strokes of different colors (see in color), (c) Medial axis segments, (d) Skeleton image.}\label{fig:testresult}
    \end{center}
    \end{figure}

There are two outputs of the algorithm.
First one is the set of vectorized medial axis segments and last one
is the final skeleton image formed by performing the extrapolation
of the medial axis segments.
In Fig.~\ref{fig:testresult}, there are four type of images: contour of the input image (Fig.~\ref{fig:testresult}(a)), contour strokes
 of different colors
(Fig.~\ref{fig:testresult}(b)), medial axis segments
(Fig.~\ref{fig:testresult}(c)), and skeleton image
(Fig.~\ref{fig:testresult}(d)).

\SubSection{Comparison of test results with other thinning methods}

\begin{figure}[t]
    \begin{center}
    \includegraphics[width=.5\textwidth,height=.45\textheight]{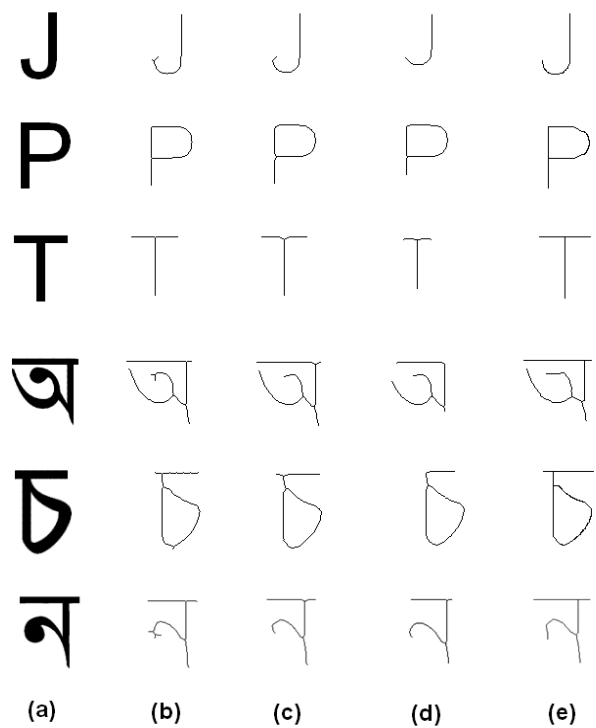}
    \caption{Comparison of results of different methods (a) Printed
             character, (b) Datta's algorithm, (c) Huang's algorithm, (d) Telea's
             algorithm, (e) Our algorithm.}\label{fig:comparisonresult}
    \end{center}
    \end{figure}

We have applied Datta's~\cite{Datta13}, Huang's~\cite{Huang01}, and
Telea's~\cite{Telea04} algorithm on printed English and Bengali
character images (Fig.~\ref{fig:comparisonresult}(a)) and compare
the results with our proposed algorithm.
We have seen that Datta's result
(Fig.~\ref{fig:comparisonresult}(b)) produces few unwanted strokes
which are not acceptable for stroke segmentation and character
recognition.
Huang's result (Fig.~\ref{fig:comparisonresult}(c)) suffers from
the distortion at the junction points of the thinned image.
Telea's result (Fig.~\ref{fig:comparisonresult}(d)) suffers from shortage of skeleton edges.
Finally, after comparing the test results of our algorithm
(Fig.~\ref{fig:comparisonresult}(e)) with all other test results, we
have seen that our method gives much better result for thinning
character images and maintains all the basic thinning properties as
well. The major improvement is that the deformation of the skeletal structure at the junction point of the skeletal branches is not
there with our results.

\Section{Conclusion}

This paper improves the performance of existent character thinning algorithms.
The main challenge of thinning character images is to preserve the shape of characters after thinning.
In spite of slight speed disadvantage, our algorithm gives quite promising result.
The proposed algorithm avoids shape distortion by detecting the ambiguous regions and avoiding the identification of medial axis points
within these regions.
The resultant skeleton maintains the pixel connectivity and is very close to the medial axis.
Additionally, our algorithm provides \textit{strokes} of character images as an intermediate result during the thinning process.
We have compared our test results with three other thinning algorithms and have observed better performance compared to all other algorithms.
The proposed thinning approach has a good potential for applications to improve the performance of OCR in Indian script.


\end{document}